%% file: main.tex
\documentclass[runningheads]{llncs}
\usepackage[T1]{fontenc}
\usepackage{verbatim}
\usepackage{graphicx}
\usepackage{hyperref}

\usepackage{algorithm}
\usepackage{algpseudocode}

\usepackage{multirow}

\usepackage{amsmath}
\usepackage{amssymb}
\usepackage{mathtools}
\usepackage{makecell}

\usepackage{pifont}
\newcommand{\cmark}{\ding{51}}

\newcommand{\smtag}[1]{^{\scriptscriptstyle\mathrm{#1}}}

\usepackage{subcaption}
\usepackage{float}
\usepackage{placeins}

\usepackage{caption}
\usepackage{xcolor}

\definecolor{shamrockgreen}{rgb}{0.0, 0.62, 0.38}
\definecolor{reviewonecolor}{rgb}{0.8, 0.3, 0.0}
\definecolor{reviewtwocolor}{rgb}{0.85, 0.2, 0.5}

\newif\ifshowme
\showmetrue 

\ifshowme
  \newcommand{\jiayi}[1]{\textcolor{shamrockgreen}{[jiayi]: #1}}
  \newcommand{\maria}[1]{\textcolor{magenta}{[MPV]: #1}}
  \newcommand{\dg}[1]{\textcolor{blue}{[DG]: #1}}
  \newcommand{\reviewone}[1]{\textcolor{reviewonecolor}{[review1]: #1}}
  \newcommand{\reviewtwo}[1]{\textcolor{reviewtwocolor}{[review2]: #1}}
\else
  \newcommand{\jiayi}[1]{}
  \newcommand{\maria}[1]{}
  \newcommand{\dg}[1]{}
  \newcommand{\reviewone}[1]{}
  \newcommand{\reviewtwo}[1]{}
\fi

\usepackage{microtype}
\usepackage{booktabs} 
\usepackage{tabularx}
\usepackage{makecell}
\begin{document}
\title{CQ4OE: A benchmark for assessing LLM-assisted ontology generation from competency questions}
\titlerunning{CQ4OE}
\author{Jiayi Li\inst{1}\orcidID{0009-0001-9475-8159} \and
Ziyuan Wang\inst{1}\orcidID{0009-0000-6228-4713} \and
Daniel Garijo\inst{1}\orcidID{0000-0003-0454-7145} \and
Mar\'ia Poveda-Villal\'on\inst{1}\orcidID{0000-0003-3587-0367}}
\authorrunning{J. Li et al.}
\institute{Ontology Engineering Group, Universidad Polit\'ecnica de Madrid 
\email{\{li.jiayi,ziyuan.wang,daniel.garijo,m.poveda\}@upm.es}\\
}
\maketitle

\begin{abstract}
Ontology generation from Competency Questions (CQs) is a central yet labor-intensive phase of Ontology Engineering. While large language models (LLMs) offer promising automation capabilities, current evaluations remain fragmented. Task formulations are heterogeneous, gold standards often lack fine-grained CQ provenance, metrics conflate lexical overlap with structural and logical adequacy, and reference ontologies are not always explicitly designed around the evaluation CQs. Here, we address these limitations with \emph{CQ4OE}, a benchmark for the systematic and reproducible evaluation of LLM-based ontology generation from CQs. For each ontology in the benchmark, we build a CQ-driven gold OWL ontology with explicit provenance linking each CQ to the classes, properties, and axioms required to answer it. From this resource, we define two complementary evaluation tasks. \emph{CQ2Term} supports term-level evaluation of CQ-specific class and property prediction over 99 CQs, and \emph{CQ2Onto} supports ontology-level evaluation over 118 CQs, including hierarchy, property modeling, and axiom-level structure.
We demonstrate \emph{CQ4OE} with experiments using nine LLMs under zero-shot, iterative, and multi-agent generation strategies, showing that LLMs recover explicit vocabulary terms more reliably than creating ontologies, particularly in property modeling, hierarchy construction, and axiom generation.

\end{abstract}

\section{Introduction}
\label{sec:introduction}

Ontology Engineering (OE) refers to the systematic process of developing machine-interpretable formal knowledge representations of a domain~\cite{salamon2022towards}. Despite well-established methodologies such as Linked Open Terms~\cite{poveda2022lot}, eXtreme Design Methodology~\cite{blomqvist2016engineering}, and
SAMOD~\cite{Peroni2016}, OE remains a
challenging activity that requires substantial expertise and
iterative cycles of requirement analysis, conceptual modeling,
implementation, and evaluation~\cite{stadlhofer2012overview,garijo2024llms}. Within these methodologies, conceptualization is a central
stage where domain requirements are transformed into an
explicit conceptual model~\cite{chavez2022chowlk}. Competency
questions (CQs) are commonly used to express these requirements
in natural language, specifying the questions that the ontology
should answer~\cite{antia2023automating}. The conceptualization step models CQs as classes,
properties, relations, and
constraints~\cite{rebboud2024benchmarking}, shaping
the structure and expressiveness of the resulting
ontology~\cite{poveda2022lot,garijo2024llms}.

Recent advances in Large Language Models (LLMs) have motivated
their adoption across ontology engineering tasks, from CQ
generation~\cite{pan2024rag,mahlaza-etal-2025-feasibility,rebboud2024benchmarking}
and ontology
conceptualization~\cite{coutinho2024leveraging,kholmska2024enhancing}
to the encoding of conceptual models into formal ontology
languages~\cite{doumanas2024integrating,masa2022ontology}.
Across these tasks, LLMs have been used to suggest candidate
terms, identify domain relationships, refine ontology fragments,
and support modeling decisions that traditionally require
substantial expertise.

However, the evaluation landscape is heterogeneous, with different studies adopting different task formulations, input/output specifications, and evaluation criteria \cite{li2025large}. This heterogeneity introduces three specific challenges. First, task formulations are frequently incompatible across studies. For example, evaluated tasks range from concept extraction and ontology completion to full OWL generation from documents or CQs~\cite{li2025large,garijo2024llms}, with each adopting different inputs, outputs, and modeling depths. Since these tasks differ fundamentally in input, output, and modeling depth, direct comparison is challenging. Second, previous analyses have shown that reference ontologies are often not finely aligned with their CQs~\cite{fernandez2021analysing}. In
particular, they rarely specify which classes, properties, or axioms are
required by each CQ, making it unclear whether a generated ontology
satisfies the CQ requirements. Third, existing metrics, often based on lexical or coarse structural overlap, do not adequately assess property modeling, logical constraints, or reasoning behavior, leaving errors such as wrong domain/range assignments, missing axioms, or flawed hierarchies
undetected. Together, these limitations make it difficult to compare LLM-based ontology generation systems under a shared, requirement-driven evaluation protocol.

To address these issues, we present \emph{CQ4OE}, a benchmark for
the systematic and reproducible evaluation of LLM-based ontology
generation from CQs. Our work makes four contributions: 
\begin{itemize}\setlength\itemsep{0.1em}

  \item \textbf{Gold standards aligned with the CQs for two evaluation tasks.}
\emph{CQ4OE} includes two complementary evaluation tasks. \emph{CQ2Term} supports term-level evaluation with CQ-to-term provenance over 99 CQs, and \emph{CQ2Onto} supports ontology-level evaluation with fine-grained CQ-to-axiom provenance over 118 CQs from six ontologies.

  \item \textbf{A multi-task evaluation framework.}
We introduce metrics for term recovery, property characteristics, domain/range triples, TBox axioms, and hierarchy closure, to assess whether a model can generate ontologies that are both structurally and logically sound beyond surface vocabulary.

  \item \textbf{An automated explainable reporting pipeline.}
We release an open-source pipeline that aligns candidate outputs with the gold standards, computes all proposed metrics, and produces detailed evaluation reports that trace matched and missing terms, axioms, CQ coverage, and reasoning-aware hierarchy recovery for each generated ontology.
\item \textbf{A baseline evaluation against CQ4OE.}
We compare nine LLMs, using three generation strategies (zero-shot term prediction for \emph{CQ2Term} and zero-shot, iterative, and multi-agent~\cite{li_maseo_2026} for \emph{CQ2Onto}).
\end{itemize}

\section{Related Work}
\label{sec:Related_work}

Large language models (LLMs) have been applied to ontology conceptualization and generation, including ontology completion~\cite{rebboud2024benchmarking}, vocabulary term suggestion~\cite{toro2024dynamic}, relation classification~\cite{babaei2023llms4ol,goyal2024silp_nlp}, and generation from user stories or CQs~\cite{perera2024exploring,saeedizade2024navigating,coutinho2024leveraging,pisu2024leveraging}. Several recent approaches directly produce ontologies from CQs. Lippolis et al.~\cite{lippolis2025assessing,lippolis2025ontology} compare prompting strategies through pitfall detection, CQ coverage, and expert assessment. MASEO~\cite{li2026maseo} generates OWL ontologies from CQs with provenance links but releases no reusable evaluation resource. These efforts remain difficult to compare because they use different CQ sets, reference ontologies, and metrics~\cite{garijo2024llms,li2025large}, and prior analysis~\cite{fernandez2021analysing,li2026maseo} shows that structural overlap with full reference ontologies is an unfair proxy for requirement satisfaction, since these ontologies may contain knowledge beyond the input CQs.

Several evaluation resources exist but target different tasks. OAEI~\cite{euzenat2011ontology} evaluates ontology matching, and BioASQ~\cite{tsatsaronis2015bioasq} evaluates biomedical question answering, neither targeting ontology construction from CQs. OntoAxiom~\cite{bakker2025ontology} identifies missing axioms within existing vocabularies instead of constructing complete ontologies. CORAL~\cite{fernandez2019coral} provides ontological requirements and CQs but does not map each CQ to the terms or axioms required to answer it. Alharbi et al.~\cite{alharbi2024characteristics} classify CQ-generation settings, focusing on CQ quality rather than the ontologies generated from them. Plu et al.~\cite{plu2024comprehensive} evaluate LLM-generated ontologies through human references and qualitative assessment, but their evaluation is not CQ-driven. Beyond ontology engineering, general LLM benchmarks such as HELM~\cite{liang2023holistic} and HumanEval~\cite{chen2021evaluating} evaluate broad capabilities, including reasoning and code synthesis, beyond ontology construction.

None of these resources provides a reusable benchmark for evaluating ontologies within a CQ-aligned
requirement scope. \emph{CQ4OE} addresses this gap with shared CQs, CQ-aligned gold standards, and metrics that assess requirement satisfaction across term recovery, property characteristics, domain/range relations, TBox axioms, and hierarchy closure.

\section{CQ4OE Benchmark Dataset Construction}
\label{sec:gold_dataset_process}

\emph{CQ4OE} is a benchmark for evaluating LLM-based ontology generation from CQs through two CQ-aligned evaluation tasks: \emph{CQ2Term} assesses whether a system predicts the classes and properties required by each selected CQ. In contrast, \emph{CQ2Onto} assesses whether a generated ontology captures the terms, property semantics, domain/range relations, axioms, and hierarchies needed to answer the selected CQs. For each source ontology and selected CQ set, we construct two task-specific gold standards. \emph{CQ2Term} records explicit terms per CQ, while \emph{CQ2Onto} records CQ-relevant terms and required axioms with CQ provenance.

\subsection{Source ontologies}
We select source ontologies based on three criteria. They must have established use in OE practice, publicly documented requirements with associated CQs, and open licenses that allow redistribution and modification. From the candidates satisfying these criteria, we select six ontologies spanning three size tiers based on the number of published CQs. The \emph{small} tier contains Wine~\cite{wine_ref} and the African Wildlife Ontology (AWO)~\cite{awo_ref}. The \emph{medium} tier contains the Open Digital Rights Language (ODRL)~\cite{odrl_ref} and
SAREF4WATR~\cite{saref4watr_ref}. The \emph{large} tier contains the Video Game Ontology (VGO)~\cite{fernandez2019coral,parkkila2017ontology} and the Software Ontology (SWO)~\cite{swo_ref}.
The selected ontologies vary in hierarchical structure, from the nearly flat ODRL (depth 1) to the deeply nested SWO (depth 15), with AWO and VGO remaining shallow but wide. This diversity lets \emph{CQ4OE} assess LLM performance across different structural complexities. Table~\ref{tab:datasets} reports key statistics for each ontology.
\input{Datasets}

\subsection{Annotation methodology}
\label{sec:annotation-methodology}

\begin{figure*}[htbp]
    \centering
    \includegraphics[width=\textwidth]{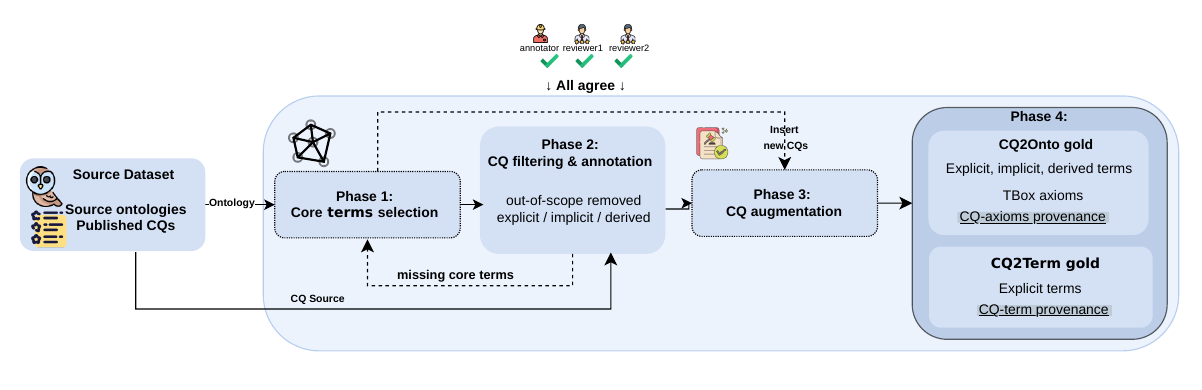}
    \caption{Overview of the CQ4OE annotation workflow. The triple-review, adjudication-based protocol applies to all candidate items across all phases, including core terms, retained CQs, term annotations, augmented CQs, CQ-to-term and CQ-to-axiom provenance links.}
    \label{fig:cq4oe_workflow}
\end{figure*}

To ensure a fair evaluation, the benchmark must assess the knowledge required by the CQs, not the broader domain content of full reference ontologies. Without CQ alignment, a generated ontology could be penalized for omitting out-of-scope content or rewarded for reproducing domain knowledge beyond the stated CQ requirements. Figure~\ref{fig:cq4oe_workflow} illustrates the four-phase workflow for constructing CQ-aligned gold standards. Table~\ref{tab:datasets} reports the resulting CQ counts per stage.

\textbf{Phase 1: Core term selection.} For each source ontology, we identify core terms (classes and properties that define the domain scope) through a two-step process combining automatic ranking with manual verification. First, we rank named classes and properties by in- and out-degree, taking highly connected nodes as initial candidates. Then, we validate these candidates with \texttt{owl2diagram}\footnote{\url{https://github.com/jatoledo/owl2diagram}} and inspect them against the official published requirements provided with the source ontologies. Terms are retained if they are highly connected, visually central, and domain-relevant. The resulting set $T_{\mathrm{core}}$ defines the conceptual backbone the gold standards must cover.

\textbf{Phase 2: CQ filtering and term annotation.} From the 255 published CQs across the six source ontologies, we remove those requiring external knowledge, unanswerable from the source ontology, or targeting instance-level rather than TBox-level knowledge. This filtering ensures fair evaluation, since every retained CQ has an answer grounded in the source TBox without requiring content that the gold ontology cannot express. 
This leaves 110 retained CQs. For each retained $\mathit{cq}_i$, we annotate three required term sets, illustrated using the AWO CQ ``Which plants eat animals?'' as a running example. $E_i$ (\emph{explicit}) contains direct surface matches, such as \texttt{Plant}, \texttt{Animal}, and \texttt{eats}. $I_i$ (\emph{implicit}) contains synonymous or equivalent phrases, such as \texttt{Carnivore} for ``predators'' in the AWO CQ ``Which animals are the predators of these animals?''. $R_i$ (\emph{derived}) contains terms not mentioned in a CQ but required for its formalization, including answer-side concepts, such as \texttt{CarnivorousPlant} for the running example.

\textbf{Phase 3: CQ augmentation.} We then check whether all core terms in $T_{\mathrm{core}}$ are covered by at least one retained CQ. Uncovered terms are addressed by manually authoring additional CQs (marked $\star$ in Table~\ref{tab:datasets}). This augmentation is conservative. Only 8 CQs (3.1\% of the original 255) are added across all domains, yielding 118 CQs in total.

\textbf{Phase 4: Gold standard construction.} With the final CQ set fixed, \emph{CQ2Term} records CQ-to-term provenance over explicit class and property terms. For \emph{CQ2Term}, we retain only CQs with at least one explicit class or property term ($E_i \neq \emptyset$), since the task evaluates explicit term recovery. This excludes 19 inference-heavy CQs from SAREF4WATR, VGO, and SWO, leaving 99 CQs. For the running example, \emph{CQ2Term} records \texttt{Plant}, \texttt{Animal}, and \texttt{eats} with CQ-to-term provenance. For \emph{CQ2Onto}, the annotated terms ($E_i \cup I_i \cup R_i$) are used to extract CQ-relevant TBox fragments from the source ontology. An extracted TBox axiom is linked to $\mathit{cq}_i$ only if its removal would make $\mathit{cq}_i$ unanswerable, so the linked axioms form a sufficient set for answering each CQ. For the running example, \emph{CQ2Onto} additionally includes the derived class \texttt{CarnivorousPlant}, together with the axioms $\texttt{CarnivorousPlant} \sqsubseteq \texttt{Plant}$ and $\texttt{CarnivorousPlant} \sqsubseteq \exists\texttt{eats}.\texttt{Animal}$ recorded with CQ-to-axiom provenance. The resulting gold standards cover 99 CQs for \emph{CQ2Term} and 118 CQs for \emph{CQ2Onto}.

Throughout construction, each annotation decision follows a triple-review, adjudication-based protocol. An annotator with an OE background produces the initial labeling of every candidate item (core terms, CQs, CQ-to-term provenance, CQ-to-axiom provenance), and two expert reviewers then inspect each item in full to verify completeness and correctness. An item is finalized only when all three reach agreement, with disagreements resolved through discussion. Most items reached agreement on first pass, with 101 of the 118 CQs (85.6\%) accepted without change. The 17 revised items concentrated in the more complex ontologies (10 in SWO, 3 in VGO, 2 in ODRL, 2 in SAREF4WATR, none in Wine or AWO), indicating that disagreements track ontology complexity rather than random variation.

\subsection{CQ2Term and CQ2Onto gold standards}
\label{sec:cq2term-cq2onto}

\emph{CQ2Term} is the term-level task. For each $\mathit{cq}_i$, the required explicit term set is $T_i\smtag{CQ2Term} = E_i$, with $T_{\mathrm{CQ2Term}} = \bigcup_i T_i\smtag{CQ2Term}$. \emph{CQ2Term} records the links between each $\mathit{cq}_i$ and its explicit terms in $T_i\smtag{CQ2Term}$ following the annotation protocol in Section~\ref{sec:annotation-methodology}. 

\emph{CQ2Onto} is the ontology-level task that preserves $O_{\mathrm{src}}$ as the domain reference ontology. For each $\mathit{cq}_i$, the CQ-relevant term set is $T_i\smtag{CQ2Onto} = E_i \cup I_i \cup R_i$, and $O_i\smtag{CQ2Onto}$ is extracted from $O_{\mathrm{src}}$ using $T_i\smtag{CQ2Onto}$ as in Section~\ref{sec:annotation-methodology}. The required TBox axioms $A_i$ are a curated subset of $(O_i\smtag{CQ2Onto})^{\mathrm{TBox}}$, obtained by first extracting all TBox axioms involving the terms in $T_i\smtag{CQ2Onto}$ and then retaining an axiom only if its removal would make $\mathit{cq}_i$ unanswerable, so that $A_i$ forms a sufficient axiom set for answering $\mathit{cq}_i$. \emph{CQ2Onto} records these CQ-to-axiom links as provenance following the same annotation protocol.




\section{Benchmark Evaluation Methodology}
\label{sec:evaluation}
We evaluate LLM outputs against the CQ-aligned gold standards described in Section~\ref{sec:gold_dataset_process}. The evaluation is organized around two principles. First, generated terms and gold-standard terms must be aligned before comparison, since models may use different labels for the same intended class or property. Second, ontology quality must be assessed at multiple modeling depths, beyond lexical overlap alone. Section~\ref{sec:term-alignment} describes the term alignment procedure, and Sections~\ref{sec:cq2term-eval} and~\ref{sec:cq2onto-eval} define the metrics for \emph{CQ2Term} and \emph{CQ2Onto}.

\subsection{Term alignment aggregation and selection}
\label{sec:term-alignment}
Generated ontologies can use labels that differ from the gold standard while denoting the same class or property (e.g., \texttt{hasPart} vs.\ \texttt{containsComponent}), so we align the gold and predicted terms before computing the \emph{CQ2Term} and \emph{CQ2Onto} metrics. From seven candidate similarity methods, we excluded WordNet-based synonym matching~\cite{sedding2004wordnet} due to poor coverage of domain terms and information-content similarity~\cite{resnik1995disambiguating,lin1998} as it requires a shared taxonomy. The final pipeline uses five methods. Hard matching applies exact string equality. Sequence matching uses Python \texttt{difflib.SequenceMatcher}\footnote{\url{https://docs.python.org/3.8/library/difflib.html}}. Levenshtein~\cite{levenshtein1966} and Jaro--Winkler similarity are computed with the \texttt{textdistance} library\footnote{\url{https://github.com/life4/textdistance}}. Embedding-based semantic similarity uses \texttt{embeddinggemma} served via Ollama\footnote{\url{https://ollama.com/library/embeddinggemma}}. Standalone Precision (P), Recall (R), and F$_1$ are reported per method, using thresholds of $1.0$ for hard matching, $0.8$~\cite{fathallahgpt} for lexical similarities, and $0.6$~\cite{rebboud2024can} for semantic similarity. Because the methods exhibit complementary failure modes, with exact and lexical matching missing paraphrases, while semantic matching can introduce false positives on short labels, the final metrics use an aggregated alignment defined below.

Let $T_g$ and $T_p$ denote the gold-standard and predicted term sets. Alignment is performed between terms of the same type. For each candidate pair $(t_g,t_p)$, where $t_g \in T_g$ and $t_p \in T_p$, each of the five similarity methods produces a score $s_m(t_g,t_p)\in[0,1]$. We aggregate the four non-hard methods using the mean of the three highest scores, with hard matching as an exact-match override:
\[
s(t_g,t_p) = \begin{cases}
1, & \text{if } s_{\mathrm{hard}}(t_g,t_p)=1,\\
\tfrac{1}{3}\sum_{m \in \operatorname{Top}_3(t_g,t_p)} s_m(t_g,t_p), & \text{otherwise,}
\end{cases}
\]
where $\operatorname{Top}_3(t_g,t_p)$ denotes the three non-hard methods with the highest scores for $(t_g,t_p)$. This aggregation mechanism prevents any single method with a lower score from vetoing a reasonable matching result.

A candidate pair passes thresholding if its score reaches the type-specific threshold $\tau$. We selected $\tau_C=0.6$ for classes and $\tau_P=0.7$ for properties by inspecting correct and spurious alignments at $\{0.5, 0.6, 0.7\}$ on a held-out subset. The higher property threshold reduces false matches on short formulaic labels (e.g., \texttt{has\_}, \texttt{is\_} prefixes), where spurious matches concentrate around $0.6$--$0.65$. Pairs passing thresholding are ranked by decreasing aggregated score, with hard matches prioritized, and selected greedily to obtain a one-to-one alignment: $(t_g,t_p)$ is kept only if both terms are still unmatched. The same procedure is applied per standalone method using its own score, enabling comparable per-method P/R/F$_1$. The accepted pairs form the final alignment set $\alpha=\alpha_C\cup\alpha_P$, where $\alpha_C$ contains accepted class pairs and $\alpha_P$ contains accepted property pairs. Since $\alpha$ is a set of one-to-one pairs, we use it in both directions to translate between gold and predicted vocabularies. This alignment provides the correspondence layer used by all downstream metrics.

Table~\ref{tab:metrics-overview} gives an overview of the metrics reported for each evaluation target. The following sections define these metrics in detail.
\input{Metrics-overview}

\subsection{CQ2Term Evaluation}

\label{sec:cq2term-eval}

\emph{CQ2Term} is the term-level task in \emph{CQ4OE}. It evaluates the explicit term
sets $T_i\smtag{CQ2Term}$ defined in
Section~\ref{sec:gold_dataset_process}. The LLM prediction for each CQ
is denoted by $\hat{T}_i\smtag{CQ2Term}$. For aggregate reporting,
let $T_g\smtag{CQ2Term}$ and $T_p\smtag{CQ2Term}$ denote the
combined gold and predicted \emph{CQ2Term} term sets:
$T_g\smtag{CQ2Term}=\bigcup_i T_i\smtag{CQ2Term}$ and $T_p\smtag{CQ2Term}=\bigcup_i \hat{T}_i\smtag{CQ2Term}$. The final alignment $\alpha=\alpha_C\cup\alpha_P$ from Section~\ref{sec:term-alignment} is applied separately for classes and properties. 


\subsubsection{Term-level metrics.}
The \emph{CQ2Term} metrics are computed globally over the combined gold term set $T_g\smtag{CQ2Term}$ and the predicted term set $T_p\smtag{CQ2Term}$. Let $\alpha$ be the final one-to-one alignment between $T_g\smtag{CQ2Term}$ and $T_p\smtag{CQ2Term}$, where $|\alpha|$ is the number of accepted matched pairs. We define true positives (TP), false positives (FP), and false negatives (FN) as $\mathrm{TP}=|\alpha|$, $\mathrm{FP}=|T_p\smtag{CQ2Term}|-|\alpha|$, and $\mathrm{FN}=|T_g\smtag{CQ2Term}|-|\alpha|$. 
Precision $P=\mathrm{TP}/(\mathrm{TP}+\mathrm{FP})$, Recall $R=\mathrm{TP}/(\mathrm{TP}+\mathrm{FN})$, and $F_1=2PR/(P+R)$ are then computed. We set $F_1=0$ when $P+R=0$. These metrics are reported per standalone similarity method and for the final aggregated alignment.

\subsubsection{CQ-conditioned coverage.}
CQ-conditioned coverage checks whether recovered terms appear under the correct CQ: a gold term for $\mathit{cq}_i$ counts as covered only if $\alpha$ aligns it to a term predicted for $\mathit{cq}_i$. Let $\mathrm{cov}_i$ be the proportion of required explicit terms covered for $\mathit{cq}_i$, and $N$ the number of retained CQs. At-least-one coverage (A1) measures the share of CQs with at least one required term recovered, mean coverage (MC) averages $\mathrm{cov}_i$ across CQs, and full coverage (FC) gives the share of CQs whose required terms are all recovered, with $\mathrm{A1}=|\{i:\mathrm{cov}_i>0\}|/N$, $\mathrm{MC}=\sum_i \mathrm{cov}_i/N$, and $\mathrm{FC}=|\{i:\mathrm{cov}_i=1\}|/N$.

The two views thus capture complementary failure modes, since a model may recover the required vocabulary without attaching each term to the CQ that requires it, leaving individual CQs unanswered despite high overall scores.

\subsection{CQ2Onto Evaluation}
\label{sec:cq2onto-eval}
\emph{CQ2Onto} evaluates LLM-generated ontologies against the ontology-level gold standard defined in Section~\ref{sec:cq2term-cq2onto} using five evaluation targets: term recovery, property characteristics, domain/range triples, TBox axioms, and hierarchy closure. For property characteristics, domain/range triples, and TBox axioms, we report a global view over the full gold and predicted element sets, and an alignment-conditioned (AC) view restricted to elements whose named terms align under $\alpha_C$ and $\alpha_P$. Term recovery is reported only globally, since term recovery itself evaluates how well $\alpha$ recovers gold terms. An AC view here would only evaluate $\alpha$ in terms that have already been aligned. Hierarchy closure is reported as a single set-based metric over inferred subsumptions translated through $\alpha$.

\subsubsection{Ontology-level metrics.}
\label{sec:ontology-level-metrics}
\textbf{Term recovery} compares predicted and gold class and property vocabularies using both the final alignment $\alpha_C\cup\alpha_P$ and the five standalone similarity methods.
\textbf{Property characteristics} compare OWL property characteristic axioms (functional, symmetric, etc.) between aligned property pairs.
\textbf{Domain/range triples} assess graph structure by
treating domain and range axioms as triples $(s,p,o)$. Properties are
compared through $\alpha_P$, class-valued domains and ranges through
$\alpha_C$, and datatype ranges by normalized string equality. Triples
involving complex anonymous OWL expressions are excluded here and
evaluated at the axiom level.
\textbf{TBox axioms} compare structural decompositions of TBox axioms, recursively translating named terms through $\alpha_C$ and $\alpha_P$ and matching datatypes, cardinalities, intersections, unions, and restrictions strictly.
\textbf{Hierarchy closure} uses HermiT to compare inferred class and property subsumption closures of the gold and predicted ontologies (with the predicted closure translated to gold vocabulary through $\alpha$), capturing hierarchy relations that may be entailed rather than explicitly asserted. For domain/range triples and TBox axioms, we additionally report an embedding-cosine diagnostic over normalized textual serializations, which does not affect strict P/R/F$_1$.

For term recovery and the non-closure structural targets, P/R/F$_1$ follow Section~\ref{sec:cq2term-eval}. For each target and view, TP counts one-to-one matched pairs that are strictly equivalent, unmatched predicted elements count as FP, unmatched gold elements count as FN, and matched but non-equivalent pairs count as both FP and FN. 

For hierarchy closure, let $\mathit{Cl}_g$ be the gold closure and $\mathit{Cl}_p^{g}$ the predicted closure translated to the gold vocabulary through $\alpha$. We define $\mathrm{TP}=|\mathit{Cl}_g\cap\mathit{Cl}_p^{g}|$, $\mathrm{FN}=|\mathit{Cl}_g|-\mathrm{TP}$, and $\mathrm{FP}=|\mathit{Cl}_p^{g}|-\mathrm{TP}+U$, where $U$ counts predicted closure pairs that cannot be translated through the alignment and are therefore counted as additional FP. P, R, and $F_1$ are computed as above.


\subsubsection{CQ-conditioned coverage.}

We report CQ coverage in \emph{CQ2Onto} in two forms: axiom-level coverage and
closure-recovered coverage. As in \emph{CQ2Term}, both are summarized using
At-least-one coverage (A1), Mean Coverage (MC), and Full Coverage (FC).
Here, coverage is computed over required TBox axioms rather than
explicit terms. Let $A_i$ denote the set of TBox axioms required to
answer $\mathit{cq}_i$, as defined in
Section~\ref{sec:cq2term-cq2onto}. For axiom-level coverage, $\mathrm{cov}_i^{\mathrm{axiom}}=\frac{\#\text{axiom-level matches in }A_i}{|A_i|}$ denotes the fraction of axioms in $A_i$ that are strictly matched by the axiom-level evaluation defined in Section~\ref{sec:ontology-level-metrics}. Closure-recovered coverage checks whether hierarchy-related gold axioms missed by axiom-level matching can be recovered through HermiT reasoning. Only missed hierarchy-related axioms are eligible for closure recovery. They are counted only if they can be decomposed into atomic \textsf{Subclass} or \textsf{Subproperty} pairs, including \textsf{EquivalentClasses} axioms whose \textsf{IntersectionOf} or \textsf{UnionOf} operands yield such pairs and all extracted pairs appear in $\mathit{Cl}_p^g$. Closure-recovered axioms are counted only among axioms not already matched by the axiom-level evaluation. The CQ coverage after adding closure-recovered axioms to the axiom-level matches is $\mathrm{cov}_i^{\mathrm{closure}}=\frac{\#\text{axiom-level matches in }A_i + \#\text{closure-recovered axioms in }A_i}{|A_i|}$. Closure recovery is restricted to gold axioms that are hierarchy-related and that decompose into atomic \textsf{Subclass} or \textsf{Subproperty} pairs. Other axioms are scored only by matching at the axiom-level.

\subsubsection{\textbf{Evaluation report.}}
\label{sec:evaluation-report}
The \emph{CQ4OE} pipeline generates a Markdown report per-run that aggregates all metrics from Sections~\ref{sec:cq2term-eval} and~\ref{sec:cq2onto-eval}, organized by evaluation target. Each report includes standalone scores for the five similarity methods, the final one-to-one term alignment, TP/FN/FP, and CQ-level traces for matched, closure-recovered, and missed axioms, together with CSV exports of term alignments and per-axiom traces. Each closure-recovered axiom is tagged as directly asserted, chain-inferred, or complex-inferred. For chain-inferred cases, a breadth-first search reconstructs the shortest path through the predicted hierarchy, providing a human-readable explanation of the recovery. Sample reports for all runs are available in the project repository.\textsuperscript{\ref{fn:repo}}

\section{Experimental Setup and Results}
\label{sec:Experimental}

We evaluate \emph{CQ2Term} on 99 CQs and \emph{CQ2Onto} on 118 CQs across six ontologies,
providing reference baselines from nine LLMs,
DeepSeek V4-Pro, V4-Flash, V3.2~\cite{deepseekv4,deepseekv32}; Qwen3.6 Plus, Flash, 35B-A3B, 27B~\cite{qwen36}; and Gemma 4 31B-IT, 26B-A4B-IT~\cite{gemma4}. Models are accessed via OpenRouter\footnote{\url{https://openrouter.ai}} with temperature $0$ and a $16{,}384$ token output limit. All setups reuse the prompting set from the MASEO Generation Agent~\cite{li_maseo_2026} so that performance differences reflect the models and strategies rather than prompt design. For \emph{CQ2Term}, it outputs the required classes and properties per CQ. For \emph{CQ2Onto}, each model generates a full ontology under three strategies. \textbf{Zero-shot} feeds the full CQ set in one pass, \textbf{iterative} feeds CQs sequentially, and \textbf{multi-agent} refines the initial ontology with RDFLib, HermiT~\cite{glimm2014hermit}, and OOPS!~\cite{poveda2014oops} for up to three iterations with backtracking. In total, \emph{CQ2Term} yields $54$ runs, and \emph{CQ2Onto} produces $162$ ontologies, providing reference baselines for future methods. All results and reports are available in the project repository.\footnote{\label{fn:repo}\url{https://github.com/oeg-upm/cq4oe-benchmark}}


\subsection{Results}
\label{sec:cq2term-results}
\begin{figure}[t!]
\centering
\setlength{\abovecaptionskip}{3pt}
\setlength{\belowcaptionskip}{0pt}

\begin{subfigure}{\linewidth}
  \centering
  \includegraphics[width=0.7\linewidth]{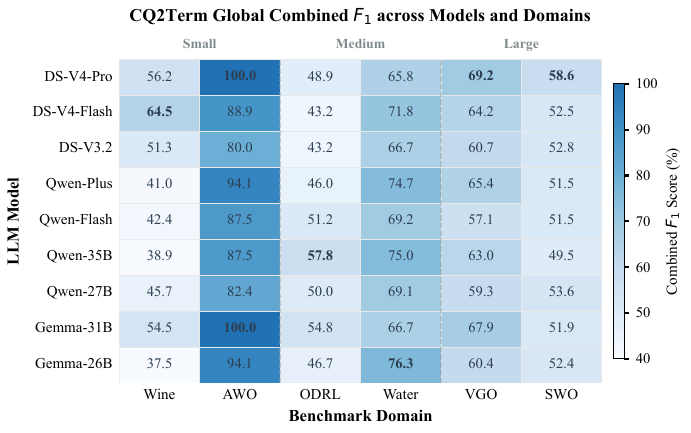}
  \caption{Overall term F$_1$ by model and domain.}
  \label{fig:cq2term_f1}
\end{subfigure}

\begin{subfigure}{\linewidth}
  \centering
  \includegraphics[width=0.7\linewidth]{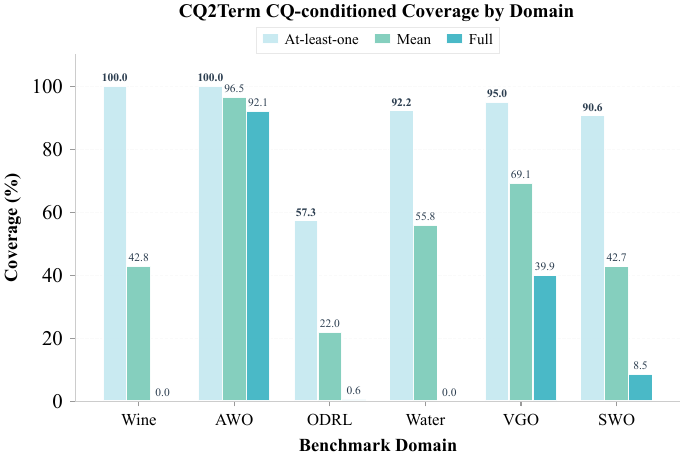}
  \caption{CQ-conditioned coverage averaged over 9 LLMs.}
  \label{fig:cq2term_cq_coverage}
\end{subfigure}

\caption{CQ2Term results across six benchmark domains.
(a) Overall term F$_1$ for each model and domain pair, computed
by pooling class and property matches before calculating
precision, recall, and F$_1$. (b) CQ-conditioned coverage
averaged over nine LLMs, reporting at-least-one, mean, and full
coverage for each domain.}
\label{fig:cq2term_results}
\end{figure}

\emph{CQ2Term} reports global F$_1$ over the combined gold and predicted term sets and CQ-conditioned coverage, in which a term counts only when recovered under the CQ that requires it. Global term F$_1$ ranges from $59.1\%$ (DeepSeek V3.2) to $66.5\%$ (DeepSeek V4-Pro), with five models leading on at least one domain. Class recovery exceeds property recovery ($67.1\%$ vs $55.2\%$), and precision-recall gaps indicate over-generation (overall $56.1\%$ vs $71.5\%$; properties $47.9\%$ vs $69.3\%$). Among standalone methods, hard matching is conservative ($51.7\%$ on classes, $30.0\%$ on properties) while semantic similarity reaches $73.9\%$ and $63.3\%$; the $33.3$-point property gap supports the multi-method aggregation in \emph{CQ4OE}. Figure~\ref{fig:cq2term_results}(a) shows global term F$_1$ for each (model, domain) pair: performance varies more by domain than by model, ranging from $48.0\%$ on Wine to $90.5\%$ on AWO. Figure~\ref{fig:cq2term_results}(b) reports CQ-conditioned coverage, averaging $89.2\%$ at-least-one, $54.8\%$ mean, and only $23.5\%$ full. Water reaches $70.6\%$ global F$_1$ but $0\%$ full coverage for every model, and Wine reaches $100\%$ at-least-one but $0\%$ full coverage. These contrasts show that strong global vocabulary recovery does not imply CQ-level completeness; by tying every recovered term to the CQ that requires it, \emph{CQ4OE} exposes requirement-localization errors that global scores hide.

\emph{CQ2Onto} extends term recovery to the full ontology structure across the five evaluation targets of Section~\ref{sec:cq2onto-eval}, under three generation strategies. DeepSeek V4-Pro achieves the highest class-label F$_1$ at $66.5\%$, and Gemma-4 31B-IT leads on property triple recovery at $41.2\%$. Mean structural F$_1$ across the 18 (domain, strategy) settings ranges from $26.7\%$ for Gemma-4 26B-A4B-IT to $33.7\%$ for DeepSeek V3.2, with five models leading on at least one domain. At the CQ level, DeepSeek V4-Flash leads on Axiom-Mean and Closure-Mean at $23.7\%$ and $24.8\%$. No single model dominates every dimension, a heterogeneity that only the multi-target design of \emph{CQ4OE} makes visible. Performance drops steeply from vocabulary to structure, with class F$_1$ averaging $59.7\%$ and property F$_1$ $31.8\%$. The AC view is consistently higher than the global view, with Triple-AC $36.4\%$ vs Triple-G $12.4\%$ and Axiom-AC $35.3\%$ vs Axiom-G $15.0\%$ (AWO excluded from triple aggregation because its gold range is a complex anonymous OWL expression). Closure F$_1$ averages only $16.7\%$, indicating shallow predicted hierarchies. Models identify relevant entities better than they assemble them into globally correct structures, a distinction that the two-view design of \emph{CQ4OE} makes explicit and that single-score benchmarks cannot capture.

\begin{figure}[t!]
\centering
\setlength{\abovecaptionskip}{2pt}
\setlength{\belowcaptionskip}{0pt}

\begin{subfigure}{\linewidth}
  \centering
  \includegraphics[width=0.7\linewidth]{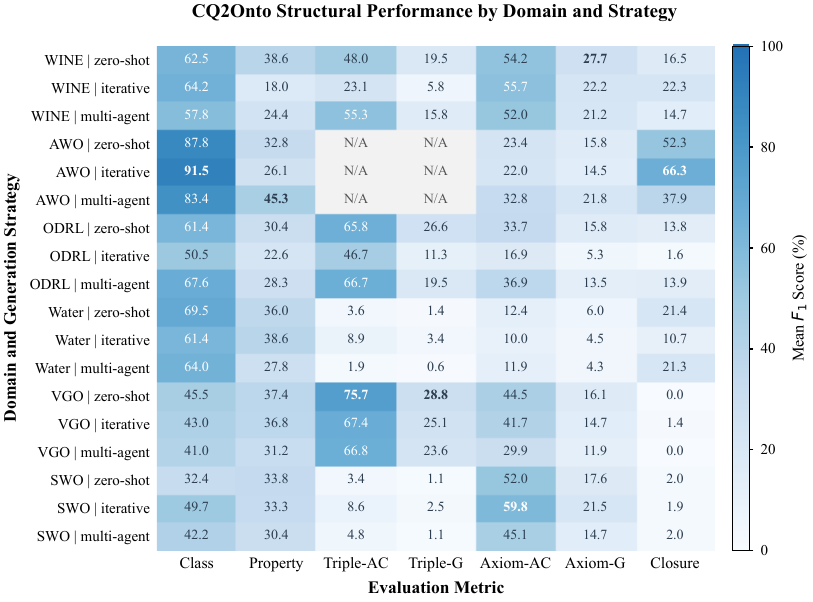}
  \caption{Structural performance (F\textsubscript{1}).}
  \label{fig:cq2onto_f1}
\end{subfigure}

\begin{subfigure}{\linewidth}
  \centering
  \includegraphics[width=0.7\linewidth]{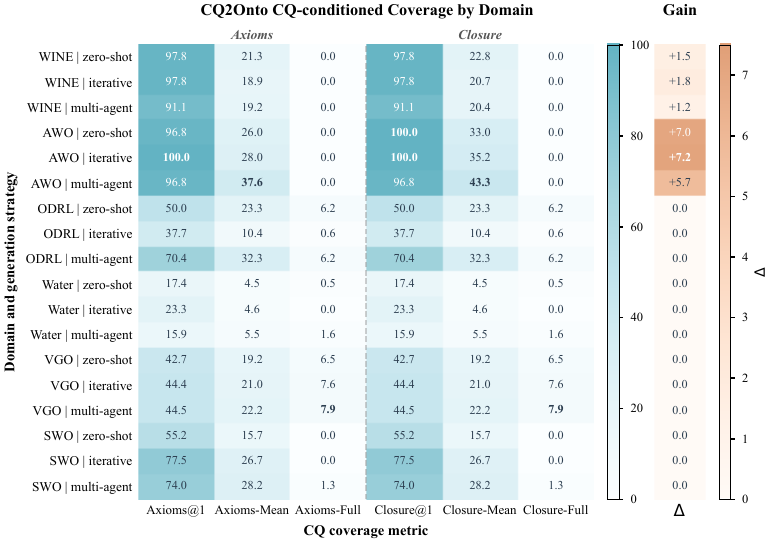}
  \caption{CQ-conditioned coverage and gain.}
  \label{fig:cq2onto_cq_coverage}
\end{subfigure}

\vspace{-0.3em}
\caption{CQ2Onto results across six benchmark domains, with each (domain, strategy) cell averaged over nine LLMs in both panels.
(a) Structural F$_1$ across seven evaluation metrics (AWO Triple cells shown as N/A because the gold range is a complex anonymous OWL expression).
(b) CQ-conditioned coverage before and after closure rescue, where $\Delta$ shows the gain in Mean coverage from closure rescue (Closure-Mean $-$ Axioms-Mean).}
\label{fig:cq2onto_results}
\end{figure}

Figure~\ref{fig:cq2onto_results}(a) summarizes the structural metrics by
(domain, strategy). Domain variation dominates strategy variation. The three
strategies yield comparable structural F$_1$ within $3$ percentage points but
shape the hierarchy differently. Iterative prompting produces the densest
\textsf{SubClassOf} graphs at $8.6$ closure pairs on average against $4.9$ for
zero-shot and $4.7$ for multi-agent. Multi-agent generation instead improves
the axioms themselves. Using OOPS! and HermiT to repair pitfalls without adding
new chains, it raises CQ-level axiom coverage (Axioms-Mean) from $18.3\%$ under
zero-shot to $24.2\%$, with the gain concentrated in AWO ($26.0\%\to37.6\%$) and ODRL ($23.3\%\to32.3\%$). Each domain exposes a different facet of the benchmark. AWO is the easiest case at class F$_1$ $87.6\%$ and closure F$_1$ $52.2\%$. Wine produces the strongest axiom-level scores in Axiom-AC $54.0\%$, Axiom-G $23.7\%$, and VGO and ODRL show large AC-to-Global drops in triples, from $70.0\%$ to $25.8\%$ and from $59.7\%$ to $19.1\%$, isolating the unaligned vocabulary as their dominant error. Water displays the sharpest split between recognition and structure, reaching class F$_1\approx 65\%$ but Triple-G $\le 3.5\%$ and Axiom-G $\le 6\%$.  Figure~\ref{fig:cq2onto_results}(b) reports CQ-conditioned coverage before and after closure rescue. Axioms@1 averages $63.0\%$, Axioms-Mean $20.2\%$, and Axioms-Full only $2.1\%$. Closure rescue raises Closure-Mean to $21.6\%$, an
average gain of $1.4$ points shown as $\Delta$ in the same panel, but leaves
Closure-Full unchanged. The gain concentrates in the domains with the densest
predicted hierarchies, where AWO rises from $30.5\%$ to $37.1\%$ and Wine from
$19.8\%$ to $21.3\%$, and there is no measurable improvement on ODRL, Water,
VGO, or SWO, whose predicted hierarchies collapse to flat edges.
\emph{CQ4OE} thus pinpoints both the failure type, recognition against
structure, and the domain factor, predicted hierarchy density against collapse,
driving each result.

Three recurrent LLM limitations emerge from these views, all traceable through the \emph{CQ4OE} reports. First, models rarely generate sufficient \textsf{SubClassOf} or \textsf{SubPropertyOf} chains. In VGO and SWO, predicted ontologies produce only $0.6$ and $0.7$ closure pairs against gold $12$ and $8$, leaving HermiT with almost no hierarchy to reason with. In Water, predicted hierarchies recover flat leaf-to-top edges (e.g., \texttt{Sensor} $\sqsubseteq$ \texttt{Asset}) and miss the chain \texttt{WaterMeter} $\sqsubseteq$ \texttt{Meter} $\sqsubseteq$ \texttt{Sensor} $\sqsubseteq$ \texttt{Device}, explaining the high class F$_1$ but near-zero closure coverage. Second, property modeling is unstable, with labels paraphrased or aliased much more often than classes, depressing property F$_1$ to nearly half of class F$_1$. Third, CQ completeness remains low, with models recovering some axioms required by a CQ but rarely all of them, leaving Axiom-Full near $2\%$. Each failure mode corresponds to a distinct evaluation target in \emph{CQ4OE} and would remain invisible under a single aggregate score.

\section{Discussion}
\label{sec:discussion}


\paragraph{\textbf{Impact and contribution.}} \emph{CQ4OE} addresses a gap in the evaluation of LLM-based ontology generation. Existing evaluations often reuse full reference ontologies as undifferentiated gold standards, even when they contain knowledge unrelated to the evaluated CQs. This makes it difficult to determine whether a generated ontology satisfies the stated requirements or merely overlaps with general domain knowledge. \emph{CQ4OE} introduces CQ-aligned gold standards in which terms and TBox axioms are explicitly linked to the CQs that require them, with provenance preserved at the level of each individual class, property and axiom. To our knowledge, no existing benchmark for ontology generation provides this level of CQ-to-axiom provenance.

This requirement-driven design advances evaluation along two complementary fronts. First, \emph{CQ2Term} targets the conceptualization capability of an LLM by measuring whether a model can recognize and organize the explicit classes and properties expressed in the CQs, which capture the most immediate semantic content of the requirements. Conceptualization is the foundation of any downstream ontology, since vocabulary selection determines all subsequent modeling decisions. By isolating this capability, \emph{CQ2Term} reduces the manual effort that ontology engineers would otherwise spend enumerating candidate terms, and offers practitioners a transparent basis for selecting the model best suited to their domain rather than relying on a single aggregate ranking. Second, \emph{CQ2Onto} evaluates the ability of a model to understand and reason over CQ requirements beyond recognizing their surface vocabulary. It assesses whether the model can move beyond the explicit terms of each CQ to recover the implicit and derived terms that the answer also requires, and whether it can express them as a coherent ontology including property characteristics, property triples, TBox axioms, and hierarchical structure under reasoner-derived closure. This goes beyond lexical fluency, requiring the structural and inferential commitments that make a CQ formally answerable. 

The transition from natural-language requirements to formal OWL representations remains a central bottleneck in ontology engineering~\cite{lippolis2025ontology,mahlaza-etal-2025-feasibility}, which makes this resource particularly relevant to the Semantic Web community. By making the CQ-to-term and CQ-to-axiom provenance explicit, \emph{CQ4OE} enables a transparent evaluation of LLM-generated ontologies, distinguishing failures that arise from missing vocabulary, unstable property modeling, shallow hierarchy generation, or incomplete coverage of the local CQs. 
The \emph{CQ4OE} pipeline produces per-run Markdown reports and CSV alignment exports that trace each metric back to specific gold and predicted axioms, making evaluation results traceable at the level of individual CQs and axioms, instead of only aggregate scores. By enabling transparent, requirement-driven comparison of ontology generation approaches, \emph{CQ4OE} can benefit not only ontology engineers but the broader Semantic Web community, as LLM-assisted ontology engineering matures and the need for fair, reproducible evaluation grows.

\paragraph{\textbf{Reusability and reproducibility.}} \emph{CQ4OE} is designed for reuse across domains, models, prompting strategies, ontology repair pipelines, and human-in-the-loop workflows. The repository organizes \emph{CQ2Term} and \emph{CQ2Onto} as two parallel directories, each with gold standards, predictions, evaluation scripts, intermediate results, and aggregated reports. To benchmark a new LLM, users add their generated ontologies to the predictions folder and run a single script that extracts atomic axioms, executes the five evaluation steps, and produces the aggregated report. To evaluate a single capability, users can apply \emph{CQ2Term} for term-level recovery or reuse the \emph{CQ2Onto} layers for ontology-level evaluation independently. To extend the benchmark, users simply need to add a new domain following the four-phase annotation methodology of Section~\ref{sec:annotation-methodology} under the same triple-review, adjudication-based protocol. Each metric has a persistent W3ID identifier and a Turtle definition in the metric catalogue, supporting integration into other evaluation pipelines. A public leaderboard\footnote{\url{https://w3id.org/cq4oe/leaderboard}} with submission guidelines is maintained in the project repository to facilitate comparison of new methods.

\paragraph{\textbf{Limitations.}} Several limitations of the current evaluation deserve attention. The most consequential is that all ontology-level metrics depend on term alignment, so errors at this layer propagate into downstream scores. Our combined hard, lexical, and semantic procedure with one-to-one selection handles lexical variation and reduces many-to-many score inflation, but it can still fail on short property labels or on semantically close yet non-equivalent terms. The thresholds $\tau_C$ and $\tau_P$ in Section~\ref{sec:term-alignment} were selected through empirical inspection, informed by previous studies~\cite{fathallahgpt,rebboud2024can}, and are applied uniformly across all six heterogeneous ontologies without per-domain tuning. The pipeline first performs a dry run that exports the complete alignment traces before any scoring. After applying the thresholds, manual inspection in all six domains confirmed that most automatic alignments agreed with expert judgment, although borderline cases remain. The thresholds are configurable, and users can inspect and, where necessary, manually correct alignments in the intermediate CSV files before re-running the evaluation.

A second limitation is methodological. The closure rescue mechanism evaluates only those hierarchy-related axioms that can be decomposed into atomic \textsf{SubClassOf} or \textsf{SubPropertyOf} relations, including \textsf{EquivalentClasses} with \textsf{IntersectionOf} or \textsf{UnionOf}. Consequently, complex expressions whose semantics cannot be represented by such atomic relations are excluded from quantitative evaluation, since assigning partial credit to individual sub-expressions within a complex axiom remains an open problem. We leave richer handling of such cases to future work.


A third concern is data contamination, since the source ontologies and some of their published CQs are public and may appear in pre-training data. Three observations qualify its impact. First, mean structural F$_1$ remains low at $26.7\%$ to $33.7\%$ across models, suggesting that prior exposure to ontology vocabulary alone does not translate into structurally correct ontology generation. Second, the CQ-to-term and CQ-to-axiom provenance is manually curated through the triple-review protocol of Section~\ref{sec:annotation-methodology}, introducing an additional manually curated annotation layer that is unlikely to have appeared in pre-training data. Third, the methodology is portable. Private or post-cutoff ontologies can be added through the documented annotation protocol, future versions of the benchmark plan will include several such ontologies to support evaluations with contamination control.

\section{Conclusion}
\label{sec:conclusion}

We presented \emph{CQ4OE}, a benchmark for evaluating LLM-based ontology generation from competency questions. \emph{CQ4OE} links each CQ to the terms and TBox axioms required to answer it through two complementary evaluation tasks, \emph{CQ2Term} and \emph{CQ2Onto}. The \emph{CQ4OE} pipeline automatically generates Markdown reports and CSV alignment exports that trace every metric back to specific gold and predicted axioms, enabling a detailed diagnosis of why a generated ontology does not satisfy individual CQs. Experiments with nine LLMs across six ontologies show that models recover explicit vocabulary more reliably than ontology structure, with performance varying more across domains than across models or generation strategies. Reasoning-based closure provides only limited recovery when the underlying hierarchy is missing. These findings highlight the need for provenance-aware, multi-dimensional evaluation of LLM-generated ontologies. We believe \emph{CQ4OE} will provide a common evaluation basis for future research on LLM-assisted ontology engineering. Future work will extend \emph{CQ4OE} with additional domains, further refine alignment and evaluation metrics, and incorporate private or post-cutoff ontologies to support contamination-controlled evaluations.

\section*{Acknowledgments.}
This work was supported by the grant ``SOEL: Supporting Ontology Engineering with Large Language Models'' (PID2023-152703NA-I00) funded by MCIN/AEI/10.13039/501100011033 and by ``ERDF/UE''.

\section*{Resource Availability Statement}
\label{sec:resource-availability}
CQ4OE is openly available under Apache 2.0 at \url{https://github.com/oeg-upm/cq4oe-benchmark}, archived on Zenodo (\url{https://doi.org/10.5281/zenodo.20080309}) and HuggingFace (\url{https://doi.org/10.57967/hf/8712}). The repository contains the CQ2Term and CQ2Onto gold standards for all six ontologies, the evaluation pipeline, and the run-level reports underlying the results. Persistent metric identifiers are defined at \url{https://w3id.org/cq4oe/metrics}.


\section*{Declaration of use of Generative AI}
All the ideas presented in this manuscript are original from the authors. During the preparation of this work, the authors used Claude (Anthropic) to improve the clarity, grammar, and readability of the manuscript. After using this tool, the authors reviewed and edited the content as needed and take full responsibility for the content of the publication.
\bibliographystyle{splncs04}

\bibliography{bib}
\end{document}

%% file: Datasets.tex
\begin{table}[htbp]
\centering
\caption{Benchmark dataset statistics. Src., Ret., and New$\star$ are original, retained, and added CQs. CQ2O and CQ2T are CQs in each gold standard. C, OP, DP, and Ax are classes, object properties, data properties, and OWL axioms. Depth is the longest SubClassOf or SubPropertyOf chain and Width is the maximum number of terms at any hierarchy level. Source counts include imports, and CQ2Onto counts refer to CQ-aligned sub-ontologies. In CQ2Term, P combines OP and DP.}
\label{tab:datasets}
\scriptsize
\setlength{\tabcolsep}{4pt}

\begin{subtable}{\linewidth}
\centering
\caption{CQ counts and source ontology statistics.}
\begin{tabular}{l rrrrr rrrrrr}
\toprule
&
\multicolumn{5}{c}{\textbf{CQs}} &
\multicolumn{6}{c}{\textbf{Source}} \\
\cmidrule(lr){2-6}
\cmidrule(lr){7-12}
\textbf{Ontology} &
Src. & Ret. & New$\star$ & CQ2O & CQ2T &
C & OP & DP & Ax & Depth & Width \\
\midrule
Wine       & 7  & 4  & 1 & 5  & 5  & 77   & 13  & 1  & 744  & 3  & 7   \\
AWO        & 14 & 7  & 0 & 7  & 7  & 31   & 5   & 0  & 93   & 2  & 21  \\
ODRL       & 35 & 13 & 6 & 19 & 19 & 30   & 49  & 4  & 416  & 1  & 22  \\
SAREF4WATR & 43 & 21 & 0 & 21 & 20 & 72   & 40  & 22 & 445  & 5  & 19  \\
VGO        & 68 & 30 & 1 & 31 & 22 & 37   & 33  & 6  & 189  & 2  & 20  \\
SWO        & 88 & 35 & 0 & 35 & 26 & 1971 & 161 & 5  & 8087 & 15 & 679 \\
\midrule
\textbf{Total} & \textbf{255} & \textbf{110} & \textbf{8} & \textbf{118} & \textbf{99} & & & & & & \\
\bottomrule
\end{tabular}
\end{subtable}

\vspace{0.6em}

\begin{subtable}{\linewidth}
\centering
\caption{CQ2Term and CQ2Onto gold standard statistics.}
\begin{tabular}{l rr rrrrrr}
\toprule
&
\multicolumn{2}{c}{\textbf{CQ2Term}} &
\multicolumn{6}{c}{\textbf{CQ2Onto}} \\
\cmidrule(lr){2-3}
\cmidrule(lr){4-9}
\textbf{Ontology} &
C & P &
C & OP & DP & Ax & Depth & Width \\
\midrule
Wine       & 11 & 5  & 17 & 8  & 1  & 68 & 2 & 4  \\
AWO        & 7  & 1  & 9  & 5  & 0  & 25 & 1 & 5  \\
ODRL       & 13 & 26 & 15 & 28  & 0  & 60 & 1 & 9  \\
SAREF4WATR & 15 & 20 & 20 & 14 & 11 & 43 & 3 & 8  \\
VGO        & 7  & 14 & 25 & 33 & 5  & 77 & 1 & 12 \\
SWO        & 20 & 18 & 31 & 18 & 5  & 80 & 1 & 7  \\
\bottomrule
\end{tabular}
\end{subtable}

\end{table}

%% file: Metrics-overview.tex
\begin{table}[tbhp]
\centering
\caption{Overview of reported metrics in CQ4OE. \emph{AC} scores use only structures whose named terms align under the alignment set $\alpha$, \emph{G} denotes global scores, computed over the full gold and predicted sets. \textbf{Per-method} reports standalone P/R/F$_1$ for the five similarity methods. \textbf{Diag.} denotes standalone embedding-cosine diagnostics. \textbf{CQ} reports at-least-one, mean, and full coverage over terms ($t$), axiom-level matches ($a$), or closure-recovered axioms ($c$).}
\label{tab:metrics-overview}
\scriptsize
\setlength{\tabcolsep}{2pt}
\begin{tabular}{l c c c c c}
\toprule
\textbf{Evaluation tasks} & \textbf{Per-method} & \textbf{Diag.} &
\textbf{P/R/F$_1$ (AC)} & \textbf{P/R/F$_1$ (G)} &
\textbf{CQ cov.} \\
\midrule
Class \& property (CQ2Term) & \cmark & ---    & ---    & \cmark & $t$ \\
Class \& property (CQ2Onto) & \cmark & ---    & ---    & \cmark & ---    \\
Property characteristics    & ---    & ---    & \cmark & \cmark & ---    \\
Triple                      & ---    & \cmark & \cmark & \cmark & ---    \\
Axiom-level                 & ---    & \cmark & \cmark & \cmark & $a$    \\
Hierarchy closure           & ---    & ---    & ---    & \cmark & $c$    \\
\bottomrule
\end{tabular}
\end{table}